\documentclass{article}

\usepackage{arxiv}

\usepackage[utf8]{inputenc} 
\usepackage[T1]{fontenc}    
\usepackage{hyperref}       
\usepackage{url}            
\usepackage{booktabs}       
\usepackage{amsfonts}       
\usepackage{nicefrac}       
\usepackage{microtype}      
\usepackage{lipsum}
\usepackage{graphicx}

\title{FSinR: an exhaustive package for feature selection}

\author{
    Francisco Arag\'on-Roy\'on\\
    Department of Computer Science\\
    and Artificial Intelligence\\
    DiCITS, DaSCI, iMUDS, \\
    University of Granada\\
    18010, Granada, Spain\\
    \texttt{far@decsai.ugr.es} \\
    \And Alfonso Jim\'enez-V\'ilchez\\
    School of Engineering Sciences\\
    DiCITS, University of C\'ordoba\\
    14071, C\'ordoba, Spain\\
    \texttt{i52jivia@uco.es} \\
    \AND Antonio Arauzo-Azofra\\
    School of Engineering Sciences\\
    DiCITS, University of C\'ordoba\\
    14071, C\'ordoba, Spain\\
    \texttt{arauzo@uco.es} \\
    \And Jos\'e Manuel Ben\'itez\\
    Department of Computer Science\\
    and Artificial Intelligence\\
    DiCITS, DaSCI, iMUDS, \\
    University of Granada\\
    18010, Granada, Spain\\
    \texttt{J.M.Benitez@decsai.ugr.es} \\
}

\begin{document}
\maketitle

\begin{abstract}
  Feature Selection (FS) is a key task in Machine Learning. It consists in selecting a number of relevant variables for the model construction or data analysis. We present the R package, FSinR, which implements a variety of widely known filter and wrapper methods, as well as search algorithms. Thus, the package provides the possibility to perform the feature selection process, which consists in the combination of a guided search on the subsets of features with the filter or wrapper methods that return an evaluation measure of those subsets. In this article, we also present some examples on the usage of the package and a comparison with other packages available in R that contain methods for feature selection.
\end{abstract}


\section{Introduction}
\label{sec:intro}


Feature selection (FS) is the process of selecting a set of relevant features (also called attributes) to build a predictive model. FS is a key part of Machine Learning and Data Science as it usually has a great impact on the final performance of the models.
Although in theory one may think that it is better to have as much information as possible, redundant, irrelevant or partially relevant features may have a negative impact on the modeling process, as the model may be learning from these features and adding noise to the result. This becomes more patent as the number of features grows.

The selection of features is a very useful tool for discovering information about the problem. Normally, no information is available about which features are useful features and which are not. FS enables the possibility of using only those features that are really important in the fitting of the models, disregarding some characteristics without affecting the solution of the problem, or frequently improving the results. In addition, the use of feature selection methods can bring other advantages such as improved interpretability, better time efficiency of model training, and reduced model complexity.

\begin{figure}[htbp]
    \includegraphics[width=1.0\textwidth]{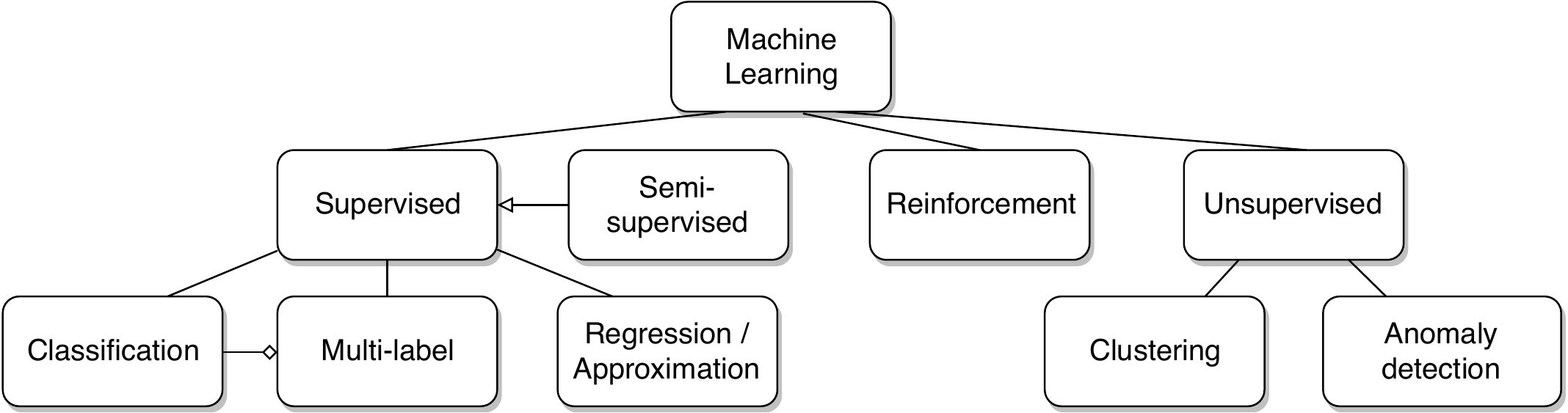}
    \caption{Machine Learning problems in which feature selection is usually applied}
    \label{fig:mlproblems}
\end{figure}

Feature selection methods are a key component of data preprocessing that can be used in most Machine Learning problems. Figure~\ref{fig:mlproblems} shows a categorization and some relations among the best known problems. As these problems are very different, FS methods are very diverse. Some methods can be applied across different problem contexts, while others are designed specifically for one category of problems.

According to the way in which FS is applied, we can identify three main classes of feature selection methods: 1) filter methods work before the learning algorithm and with complete independence, 2) wrapper methods use the learning algorithm to evaluate the relevance of features, and 3) embedded methods include FS inside the learning algorithm.

In spite of the vast amount of methods developed, keeping an exhaustive list of feature selection methods, FS is an open problem for which no definitive solution has been found and actually no single solution is expected. One of the possible reasons is the complexity of the FS problem, which leads to the suspicion that there is no universal solution to all contexts. This is because FS is an NP-Complete problem. An NP-Complete problem is a type of computational problem for which no efficient solution to every instance has been found and it is expected that none will be found. The reason is that solving it requires a time that is exponential at least to the size of the problem, which makes the execution time to grow faster than the size of the problem. That is why it is interesting and justified the regular appearance of contributions with new proposals to apply FS in new areas and scopes. Nowadays the number of published papers describing new FS methods are numbered in hundreds.


In the Comprehensive R Archive Network (CRAN) repository at \url{http://CRAN.R-project.org/}, many packages implementing FS methods can be found. Some of them are packages that have many functions dedicated exclusively to this task, such as the \textbf{FSelector} \cite{FSelector18} and the \textbf{MXM} package \cite{MXM17}. \textbf{FSelector} contains several wrapper and filter methods, as well as cutoff methods for choosing a subset based on the weight of the features, and the \textbf{MXM} package also includes a large number of methods for FS and addresses the problem in depth, but this package contains methods less known because the methods are based on a series of specific algorithms mostly proposed by them. The rest of the packages address the problem of FS in lesser depth and shorter extent. Packages such as \textbf{Boruta} \cite{Boruta18} and \textbf{spSFR} \cite{spFSR18} provide only one particular method specifically implemented to carry out feature selection while others such as \textbf{varSelRF} \cite{varSelRF07} and \textbf{featurefinder} \cite{featurefinder18} contain some methods for making tree-based feature selection. The \textbf{EFS} \cite{EFS17} package offers a different approach by proposing an ensemble of feature selection methods. On the other hand, packages such as \textbf{caret} \cite{Caret08}, \textbf{CORElearn} \cite{CORE18}, \textbf{RWeka} \cite{Weka09} or \textbf{mlr} \cite{MLR16} also contain feature selection methods. But these packages are not dedicated to feature selection, and therefore the development of FS methods is limited.


The \textbf{FSinR} package, presented in this article, aims to provide as complete a tool as possible for feature selection. The package not only aims to offer many methods of feature selection but also aims to offer the most widely used methods in literature. Unlike the previous packages, the \textbf{FSinR} package offers great versatility in using and combining filter and wrapper methods with search methods. In addition, we use as wrapper measures the methods available in the \textbf{caret} package, offering a solid background for operation and a wide variety of methods. Regarding to the use of the package, it consists of applying theses filter and wrapper measures on a dataset, and for this, search methods can be applied to find the best possible subset of features. The package is available for use in the CRAN repository at: \url{https://cran.r-project.org/package=FSinR}


The remainder of the article is structured as follows. Section 2 provides an overview of feature selection theory. Section 3 describes the design of the package and its internal details. Section 4 explains its usage providing real case examples. Section 5 compares with the previous packages available in CRAN. Section 6 explains the contribution of the proposed package. Finally, section 7 summarizes the conclusions of this paper. 

\section{Feature Selection}

\begin{figure}[htbp]
\centering
\includegraphics[width=.9\textwidth]{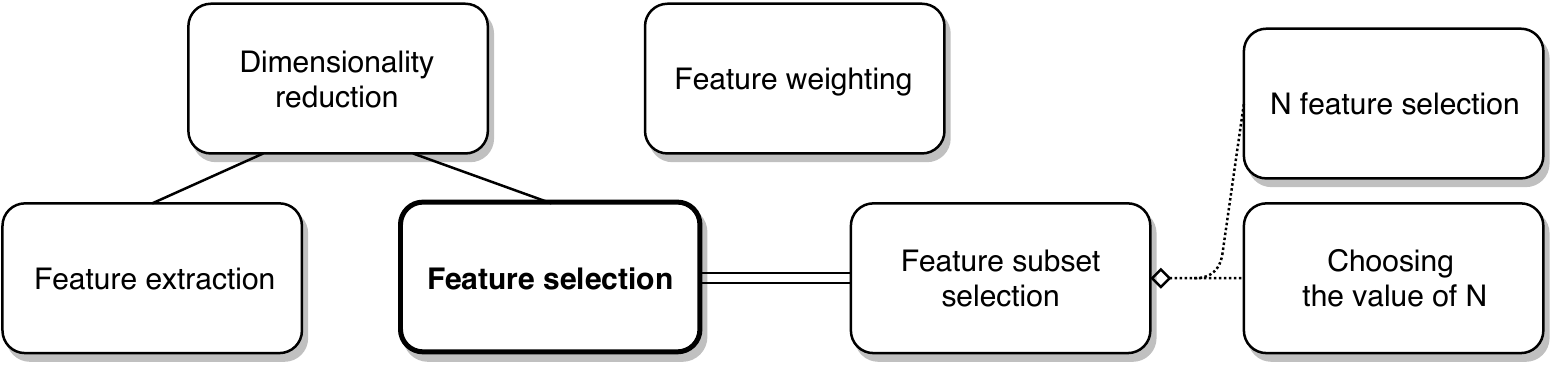}
\caption{Context of feature selection and its related problems}
\label{fig:fsrelatedprob}
\end{figure}

Feature Selection (FS) is part of the preprocessing techniques~\cite{garcia2015data}, steps that ease the learning process from data sets. Among them, FS belongs to the group of \textit{dimensionality reduction} tools whose aim is to reduce the size of the data. Aside from FS, the problem of choosing a set of features is also known as variable, attribute, or predictor selection. There are also other related problems, shown in Figure~\ref{fig:fsrelatedprob}. This introduction clarifies their relation with FS and how this is important for the implementation of FS methods.

\textit{Feature extraction} is another type of dimensionality reduction techniques. These techniques achieve the reduction by creating new combinations of condensed features, trying to summarize all the information in a reduced set of new features. This should not be confused with FS, as FS use the original features unmodified, choosing the features that are useful for solving the learning problem and discarding the others (a subset of the available features). Which technique is more adequate depends on the context of the problem. For example, when mixing different feature types or if interpretability of the developed model is important, FS is probably a better choice.

\textit{Feature weighting} assigns a weight to each feature in a data set. Many FS methods use some feature weighting technique to evaluate features. However, this is not feature selection in a narrow sense (some criterion is needed in addition to the evaluation) and it can be used for other purposes. For example, feature weighting combines very well with k-nearest-neighbor (knn) learners.

The term \textit{feature subset selection} is used to refer to those FS methods that work with sets of features rather than individual features. As an alternative approach rather than searching for a subset of features inside the power set of all features, \textit{N feature selection} is a more specific problem, finding the (\textit{best}) subset with a given number (\textit{N}) of features. This is a sensible idea, as we may know the maximum amount of features a learner support and we may not be interested in removing other features (for example, if we suspect all of them can contain valuable information or we know the learner handles noise well).

Finally, we can consider the problem of automatically \textit{choosing the value of N}, which, together with N feature selection, represents another way of approaching feature selection.

\subsection{Characterization of feature selection methods}

\begin{figure}[htbp]
\centering
\includegraphics[width=.7\textwidth]{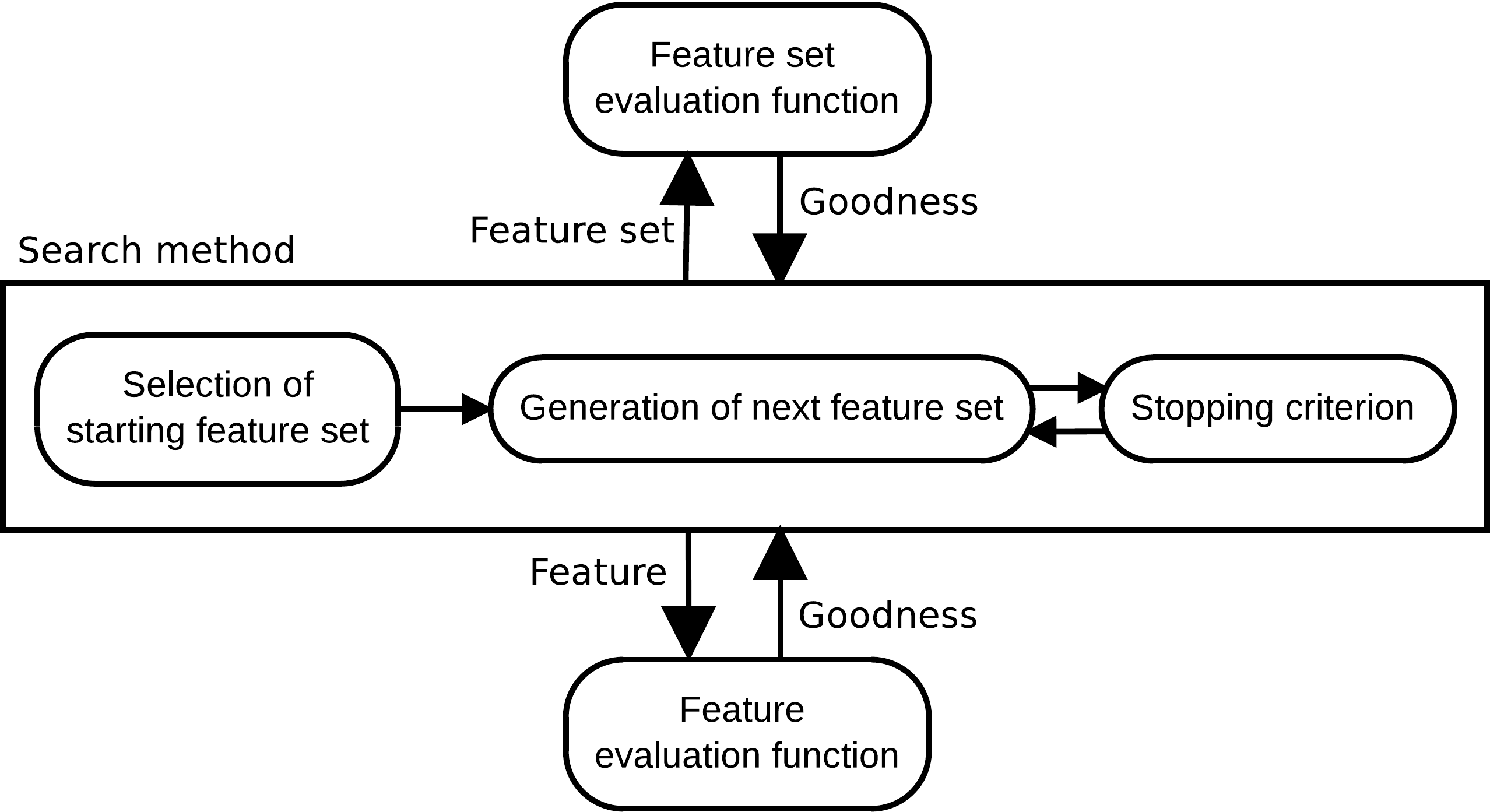}
\caption{Modularization of feature selection methods}
\label{fig:fsprocess}
\end{figure}

As all feature selection methods end up providing a set belonging to the powerset of all features, they can all be seen as a search in this space. Besides, all methods evaluate features or sets of features somehow. Consequently, the search strategy guides the search according to the evaluation inside this large search space ($2^n$, where $n$ is the number of features). For this reason, in order to characterize all methods, we will use the modularization of feature selection methods~\cite{ArauzoAzofra2011empirical} that is shown in Figure~\ref{fig:fsprocess}. Thus the two main components are the evaluation function and the search method.

\subsection{Classification of the feature selection methods by evaluation approach}
\label{subsec:classification_feature_selection_methods}

One of the most widely accepted forms of classification of feature selection methods is by how they are applied, which is directly related with how they evaluate features. Three main approaches are identified.

\subsubsection{Embedded Methods}

In some learning algorithms, the feature selection is implicitly implemented along with the learning process. These FS methods have the advantage that being specifically designed, their performance is expected to be higher. Some well-known examples of embedded methods are the RIDGE and LASSO algorithms, as well as decision trees.

\subsubsection{Wrapper Methods}

Wrapper methods use the learning algorithm to evaluate the quality of feature subsets. The main advantage of this approach is that the feature evaluation is performed in the real environment in which it is going to be applied, and therefore, it takes into account the particularities of the learning algorithm. However, these methods are computationally very expensive.

\subsubsection{Filter Methods}

Filter methods evaluate features based on a utility measure. These methods first apply the feature selection algorithm. Then, the learning method is applied only to the previously selected features. Therefore, these methods have the advantage of being independent of the learning algorithm and they can be used with anyone regardless of their computational efficiency or limitations. This is possibly the most frequently used approach.

\begin{figure}[htbp]
\centering
\includegraphics[width=1.0\textwidth]{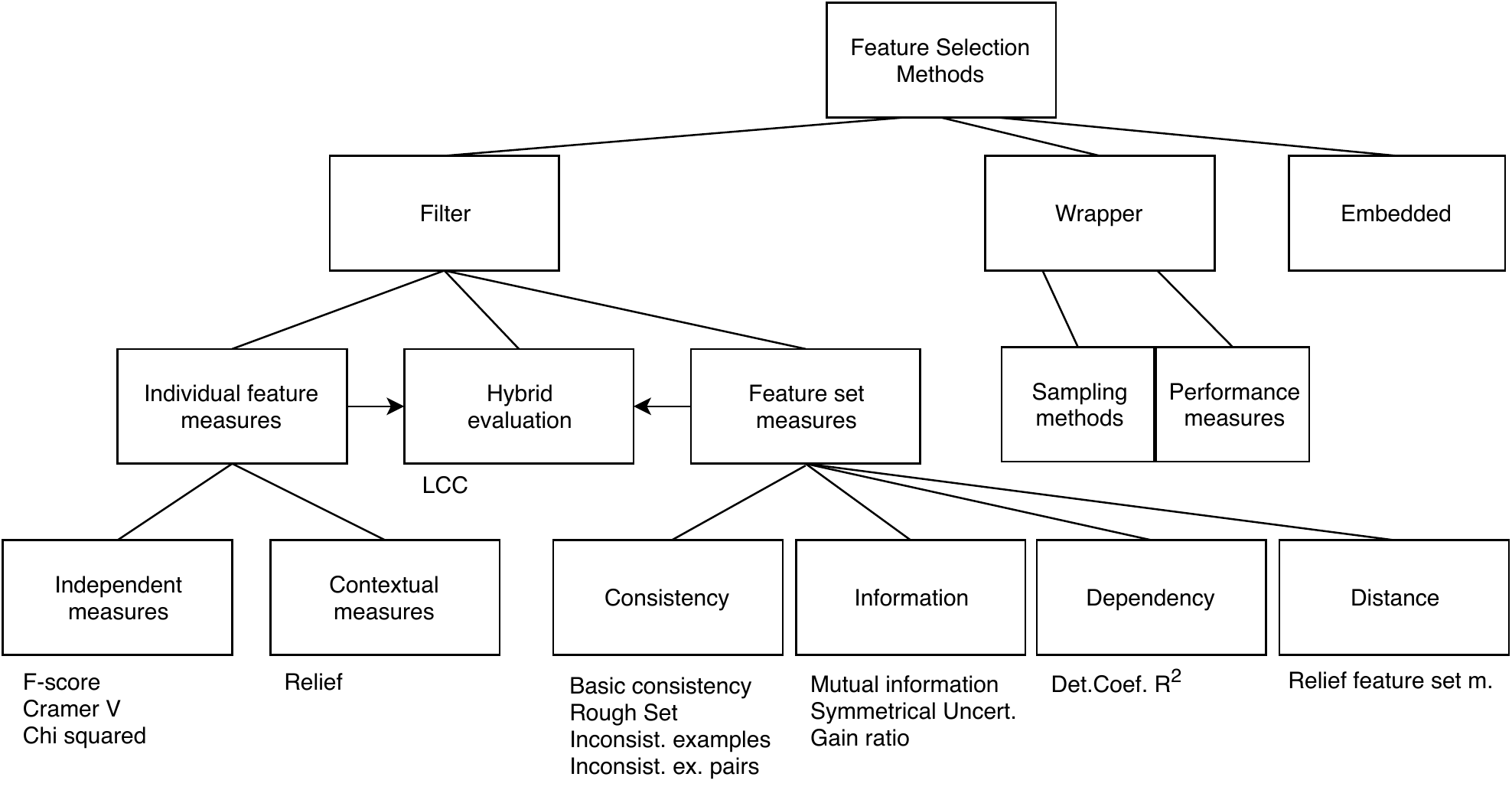}
\caption{Classification of FS methods according to the feature evaluation approach}
\label{fig:fsmeval}
\end{figure}

\paragraph{Utility measures}

In filter methods, it is necessary to measure the suitability of the features. There are two main types of approaches to measure the utility of features, individually and in sets. The main advantage of an individual feature evaluation function is its simplicity. Two subtypes of individual measures can be identified. The most common, independent individual measures, use no information from the other features, just the feature evaluated and the response variable. Examples of this measures are: F-score, Cramer V or Chi squared. On the contrary, Relief derived measures are contextual, because they also consider the distance among examples with the values of all the other features. This is probably the reason for its success.


The other main type of approaches is using feature set evaluation functions. Their advantage is the ability to detect combined interactions with the response variable. This is a more accurate approach to reality, where interactions are complex. Nevertheless, evaluating all subsets ($2^n$) is a much more computationally expensive process than evaluating $n$ individual features. As evaluating all sets is not usually plausible, heuristic searches are applied and the advantage of the more advanced evaluation gets downgraded.

Finally, there are hybrid FS methods combining both approaches called hybrid methods. For example, LCC~\cite{ShinXu2009} first uses an individual measure to sort features. Then, this order of features guides the search that evaluate feature sets with a consistency feature set measure.


\subsection{Classification of the feature selection methods by search strategy}
\label{subsec:search_methods}

\begin{figure}[htbp]
\centering
\includegraphics[width=1.0\textwidth]{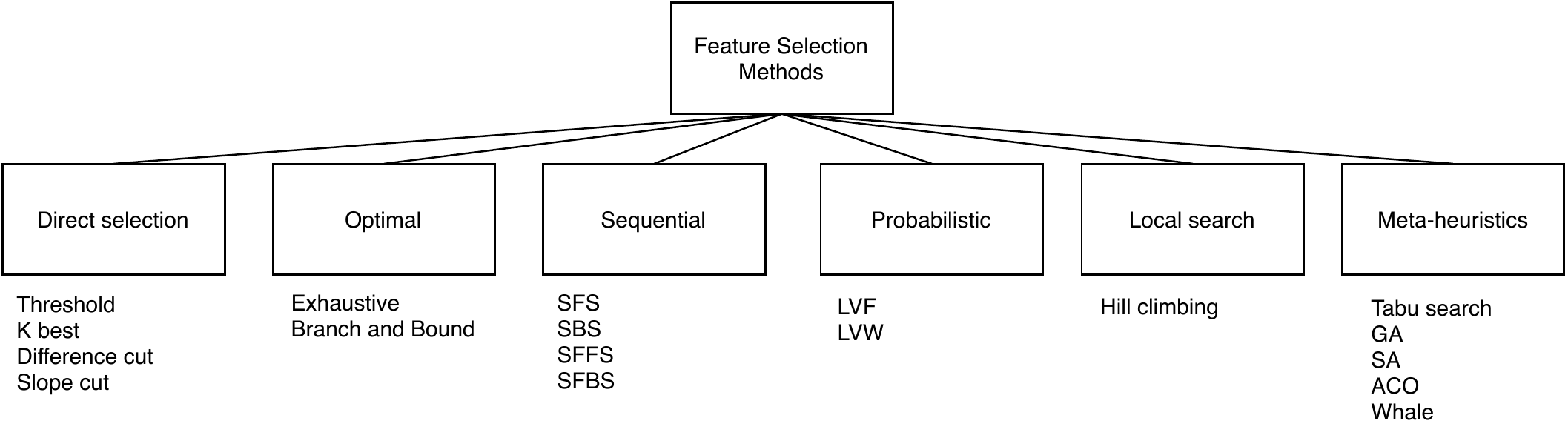}
\caption{Classification of FS methods according to the search strategy}
\label{fig:fsmsearch}
\end{figure}

Seeing feature selection as a search problem, several types of search strategies can be identified, as illustrated in figure~\ref{fig:fsmsearch}.

\subsubsection{Direct selection}

These are not really a searches. They select one set directly from the search space based on individual evaluation measures, for example, selecting all features whose evaluation is over a threshold, or the $K$ highest valued features.

\subsubsection{Optimal search}

These search strategies guarantee to find the optimal set of features. The most evident example is the exhaustive search, which evaluates all possible subsets to keep the best, but also Branch and Bound search provides this guarantee when used with measures that satisfy the monotonic property. Since FS is an NP-Hard problem, this strategy can only be applied to instances of small sizes.

\subsubsection{Sequential search}

These search strategies are characterized by taking a path and never going back in a greedy approach. They are the simplest search strategies. The most common sequential search strategies are the SFS (Sequential Forward Selection) that starts with the empty set of features and adds the feature that most improve the performance of the selection, and the SBS (Sequential Backward Selection), which performs the search in the reverse direction, removing the feature that less degrade performance.

\subsubsection{Probabilistic search}

Probabilistic algorithms are those that leave some of their decisions to randomness. These algorithms randomly explore the search space. An example is the Las Vegas wrapper algorithm, which in a random way explores subsets of features while retaining the best, to finalize after a certain number of iterations.

\subsubsection{Local search}

Local search is an iterative process that begins with an initial solution and searches your neighborhood for a better solution. If the method finds a better solution, it replaces its current solution with the new one and continues with the process until a certain stop criterion is met. Local search is the basis of many of the methods used in optimization problems.

\subsubsection{Metaheuristics}

Metaheuristics are techniques designed to solve complex problems. These techniques intelligently combine different concepts to adequately explore the search space. Among the most relevant metaheuristics can be emphasized: a) genetic algorithms, which are based on biological evolution, and which employ genetic operators existing in nature such as crossing or mutation, b) simulated annealing, based on concepts of statistical mechanics, more specifically the cooling of a solid, and c) taboo search, in which a memory structure is maintained that prevents the algorithm from revisiting previously visited solutions.

\section{Package architecture and implementation details}


Package \textbf{FSinR} is written in the language \textbf{R}. Feature Selection methods implementaed in \textbf{FSinR} are created by combining a search algorithm, whose goal is to find the best set of features to use, and one or more measures to assign a score to each feature or set of features.

\begin{figure}[htbp]
\includegraphics[width=1.0\textwidth]{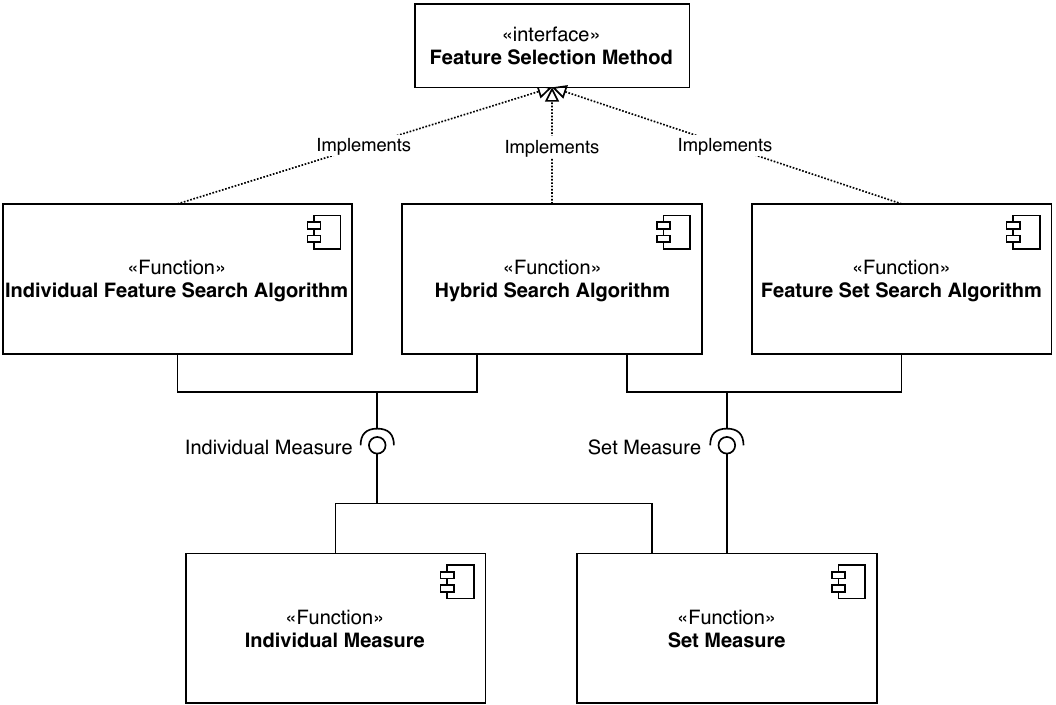}
\caption{FSinR structure}
\label{fig:FSinRstructure}
\end{figure}

As detailed in Figure \ref{fig:FSinRstructure}, FSinR provides three kinds of search algorithms: individual feature, feature set and hybrid, depending on how they evaluate features (hybrid algorithms use both individual and set measures during its process).

FSinR also provides individual feature measures and feature set measures, which can be used as individual feature measures as well, as one feature can be considered a single element set. These measures share a common interface, so they are as interexchangeable as possible.

The package provides 17 measures (listed in Table \ref{table:measures}), as well as 20 different search strategies (listed in Table \ref{table:searchAlgorithms}) that can be combined (Figure \ref{fig:FSinRstructure}) to create a wide variety of feature selection methods. Moreover, wrapper evaluation measure is compatible with the 238 learning models for both classification and regression available in the \textbf{caret} package. Most of these combinations corresponds to previously proposed method. The package provides a general framework within which most proposals can be fitted.

\begin{table}
\begin{center}
\resizebox{\textwidth}{!}{%
\begin{tabular}{|l|l|l|l|l|}
\hline
Method & Approach & Property & Evaluation & Reference \\
\hline \hline
Chi-squared & Filter & Dependency & Individual & \cite{Pearson1900} \\
Cramer V & Filter & Dependency & Individual & \cite{Cramer1946} \\
F-score & Filter & Distance & Individual & \cite{Wang2018} \\
Relief & Filter & Distance & Individual & \cite{Kira1992} \\
Rough Sets consistency & Filter & Consistency & Set & \cite{Pawlak1982} \\
Binary consistency & Filter & Consistency & Set & \cite{AlmuallimDietterich1991} \\
Inconsistent Examples consistency & Filter & Consistency & Set & \cite{DashLiu2003} \\
Inconsistent Examples Pairs consistency & Filter & Consistency & Set & \cite{Arauzo2007} \\
Determination Coefficient ($R^2$, to continous features) & Filter & Dependency & Set & \cite{rsquared} \\
Mutual information & Filter & Information & Set & \cite{QianShu2015} \\
Gain ratio & Filter & Information & Set & \cite{Quinlan1986} \\
Symmetrical uncertainty & Filter & Information & Set & \cite{WittenFrank2005} \\
Gini index & Filter & Information & Set & \cite{Ceriani2012} \\
Jd evaluation & Filter & Information & Set & \cite{Narendra1977} \\
MDLC evaluation & Filter & Information & Set & \cite{Sheinvald1990} \\
RFSM evaluation & Filter & Distance & Set & \cite{Arauzo2004} \\
Wrapper function & Wrapper & Performance & Set & \cite{kohavi1997} \\ \hline
\end{tabular}}
\caption{Feature measures implemented in \textbf{FSinR}}
\label{table:measures}
\end{center}
\end{table}

\begin{table}
\begin{center}
\resizebox{\textwidth}{!}{%
\begin{tabular}{|l|l|l|l|}
\hline
Method & Evaluation & Approach & Reference \\
\hline \hline
Breadth First Search & Set & Optimal & \cite{Kozen1992} \\
Deep First Search & Set & Optimal & \cite{Kozen1992} \\
Hill-Climbing & Set & Local search & \cite{Russell2009} \\
Tabu search & Set & Meta-heuristics & \cite{glover1986} \cite{glover1990} \\
Genetic algorithm & Set & Meta-heuristics  & \cite{yang1998feature} \\
Whale optimization algorithm & Set & Meta-heuristics  & \cite{Kumar2018} \\
Simmulated annealing & Set & Meta-heuristics  & \cite{KirkpatrickGelattVecchi1983} \\
Ant colony optimization & Set & Meta-heuristics & \cite{Kashef2015}  \\
Las Vegas wrapper & Set & Probabilistic & \cite{LiuSetiono1996}  \\
Sequential Forward Selection & Set & Sequential & \cite{Whitney1971}  \\
Sequential Floating Forward Selection & Set & Sequential & \cite{Pudil1994}  \\
Sequential Backward Selection & Set & Sequential & \cite{MarillGreen1963}  \\
Sequential Floating Backward Selection & Set & Sequential & \cite{Pudil1994}  \\
Linear Consistency-Constrained & Hybrid & Sequential &  \cite{ShinXu2009} \\
Select K best & Individual & Direct selection & \cite{ArauzoAzofra2011empirical} \\
Select Percentile & Individual & Direct selection & = \\
Select threshold & Individual & Direct selection & = \\
Select threshold range & Individual & Direct selection & = \\
Select difference & Individual & Direct selection & = \\
Select slope & Individual & Direct selection & = \\ \hline
\end{tabular}}
\caption{Search strategies implemented in \textbf{FSinR}}
\label{table:searchAlgorithms}
\end{center}
\end{table}

\subsection{Measures}

Utility measures described at section~\ref{subsec:classification_feature_selection_methods}
 are implemented differently according to the quantity of features they evaluate.

\subsubsection{Measures on individual features}

Individual feature measures are implemented following a common functional interface:

\begin{verbatim}
function(data, class, features, params...)
\end{verbatim}

The parameter \texttt{data} contains the data set, the parameter \texttt{class} is a string containing the class name and the parameter \texttt{features} is the feature name represented by a string. If a vector of feature names is passed, every measure will evaluate independently.

Depending on the measure, extra parameters may be available. They are configured with default values, though, so these parameters are optional.

These functions return a numeric value. In case of evaluating multiple features, they return a vector containing the value for each feature.

In addition, the functions have three attributes: name, maximize (true if results are better when the value is higher, false otherwise) and type (individual measure, set measure...)

\subsubsection{Measures on feature sets}

Feature set measures are implemented following a common functional interface:

\begin{verbatim}
function(data, class, features, params...)
\end{verbatim}

Compared to individual measures, the parameters are the same, but feature set measures always returns a numeric value, as they evaluate the vector of features as a set.

The same attributes as individual features are also shared.

\subsubsection{Wrapper Generator}

This package allows to measure feature sets using wrapper measures, as mentioned in section~\ref{subsec:classification_feature_selection_methods}.

They are created by means of the \texttt{wrapperGenerator} function:

\begin{verbatim}
wrapperGenerator(learner, resamplingParams, fittingParams)
\end{verbatim}

The parameter learner expects the learner to be used and it is required. \texttt{resamplingParams} and \texttt{fittingParams} are the control parameters for evaluating the impact of model tuning parameters and control parameters for choose the best model across the parameters respectively.

\texttt{resamplingParams} arguments are the same as those of the \textbf{caret} \texttt{trainControl} function and \texttt{fittingParams} arguments are the same as those of the \textbf{caret} \texttt{train} function (minus the parameters: \texttt{x}, \texttt{y}, \texttt{form}, \texttt{data}, \texttt{method} and \texttt{trainControl}).

This function returns a wrapper function to be used as a measure function. It shares the common functional interface and attributes with the rest of feature set measures.

\subsection{Search algorithms}

Search algorithms described at section~\ref{subsec:search_methods} are implemented differently according to how they work with features.

\subsubsection{Feature set search algorithms}

Feature set search algorithms are implemented following a common functional interface:
\begin{verbatim}
function(data, class, featureSetEval, params...)
\end{verbatim}

The \texttt{featureSetEval} parameter expects the feature set measure function used to evaluate features. If an individual measure is used, an error will be shown. The rest of parameters are the same as in evaluation measures.

They return at least a list with two elements: \texttt{bestFeatures}, a vector containing ones and zeroes to represent which features have been selected or not respectively, and \texttt{bestValue}, its evaluation score. This list may additionally contain another set of elements representing information from the running of the search algorithm. These elements are different for each algorithm.

\subsubsection{Individual feature search algorithms}

Individual feature search algorithms are implemented following a common functional interface:
\begin{verbatim}
function(data, class, featureEval, params...)
\end{verbatim}

The \texttt{featureEval} parameter expects the feature measure function used to evaluate features. Both individual measures and set measures can be used. The rest of parameters are the same as in feature set search algorithms.

In addition to \texttt{bestFeatures}, described in feature set search algorithms, individual feature search algorithms also return two elements, \texttt{featuresSelected} (the names of the returned features sorted according to the result of the evaluation measure) and \texttt{valuePerFeature} (the evaluation measures of the returned features).

\subsubsection{Hybrid search algorithms}

Hybrid search algorithms are implemented following a common functional interface:
\begin{verbatim}
function(data, class, featureSetEval, featureEval, params...)
\end{verbatim}

Their parameters are the same as the ones explained in previous search algorithms.

They also return the same elements as feature set search algorithms.

\section{Using the package}

In this section, we illustrate the use of the package. More detailed information regarding \textbf{FSinR} package usage as well as additional and more detailed examples can be found at the package website at \url{https://dicits.ugr.es/software/FSinR/}. The main use of the package is to carry out a feature selection on regression or classification problems guided by a search algorithm. This search algorithm selects subsets of features and evaluates them according to the measure returned by the filter or wrapper methods. The filter and wrapper methods can also be used individually, without a search algorithm, or with a direct selection method. For the sake of brevity we summarize here two simple examples. 

We show how to apply feature selection to a dataset, and the procedure, functions, and parameters required for this task. In the following examples, we demonstrate the use of filter and wrapper measures using search methods. For the wrapper case, we use a specific search method, Tabu search (\texttt{ts()}), and a specific learning algorithm, knn, to solve a classification problem. For filter, we use a genetic algorithm with the filter measure, Gini Index, as a fitness measure. For these examples we use \texttt{wine} and \texttt{iris} datasets, which are benchmarking datasets illustrating classification problems. Of course we intend to maximize the evaluation measure.
The \texttt{wine} dataset can be downloaded from the UCI Machine Learning Repository at \url{https://archive.ics.uci.edu/ml/datasets/Wine}, and the \texttt{iris} dataset can be obtained directly in \textbf{R}. We note that the \textbf{FSinR} package may contain different parameters depending on the method used to build it, and therefore, the descriptions presented may be different when using other methods.

\subsection{Wrapper approach}

The following is an example of the use of the package that performs a feature selection of the \texttt{wine} dataset using a knn as a wrapper method utilizing the tabu search as the search algorithm.

The first step is the wrapper method preparation. The function that generates the wrapper method is \texttt{wrapperGenerator()}. This function serves as an interface to the \texttt{train()} and \texttt{trainControl()} functions of the \textbf{caret} package and is therefore intended as a container for these functions. The \texttt{wrapperGenerator} function contains three parameters that must be prepared for the correct operation of the learning method. The first of them, \texttt{learner}, is the name of the learning method to be used. It must be a method supported by the \textbf{caret} package. In this case the method used in the example is ``knn''. The second parameter, \texttt{resamplingParams}, is a list that determines the evaluation of the model, and therefore the parameters that can be set are the same as those that can be set in the \texttt{caret} \texttt{ainControl()} function. In this case, we choose a 10 fold cross-validation. The third parameter is \texttt{fittingParams}, which is a list that determines the parameters for fitting the model. \texttt{fittingParams} will contain the same elements as the \texttt{train()} function of \textbf{caret}, except for the parameters relating to the dataset, method and training (x, y, form, data, method and trainControl). For this parameter, we choose to carry out a preprocessing of the data consisting of a centering and a scaling, as well as to select the accuracy as the error measures and to establish a grid of parameters for the knn method.

\begin{verbatim}
R> resamplingParams <- list(method='cv', number=10)
R> fittingParams <- list(preProc=c('center','scale'), metric='Accuracy',
+   tuneGrid=expand.grid(k=1:20)))
R> wrapper <- wrappergenerator('knn', resamplingParams, fittingParams)
\end{verbatim}

The wrapper method generated can be used directly to obtain a measure of the subset of features passed to it as a parameter. To do this, the dataset, the target feature and the subset of features are passed as parameters.

\begin{verbatim}
R> wrapper(wine,'V14',c('V1','V2','V3','V4','V5','V6','V7','V8','V9','V10','V11','V12',
+   'V13'))
\end{verbatim}

The second step is to establish the search strategy and integrate the wrapper measures into the search process, and then perform the feature selection. As a general rule, all search methods need the same three inputs/parameters to run. The rest of the parameters are specific to each function and are established by default, although it is interesting to define them in order to obtain an optimal result. The three mandatory parameters are \texttt{data}, \texttt{class} and \texttt{featuresSetEval}. First, the \texttt{data} parameter represents the dataset on which feature selection is performed. This parameter must be a data.frame or a matrix ($m \times n$) where $m$ is the number of instances and $n$ is the number of features. The \texttt{class} parameter represents the name of the column that contains the output feature. The package internally determines, according to this column, whether the problem is a regression (numerical variable) or a classification (categorical variable) problem. Finally, the parameter \texttt{featureSetEval} represents the method that returns the evaluation measure of the features. For the case of this example, we choose tabu search, \texttt{ts()}, as the search strategy and the input parameters are 1) the dataset \texttt{wine}, 2) the parameter \texttt{class} is ``V14'' and 3) the wrapper method generated in the previous step. The default function call is as follows:

\begin{verbatim}
R> result.search.fs <- ts(wine, 'V14', wrapper)
\end{verbatim}

In this example, some of the default parameters are modified. More specifically, we set the size of the tabu list to 4, the number of tabu search iterations to 10, and an intensification phase and a diversification phase, both of 5 iterations. Finally, to visualise the execution of the algorithm on the console, we set the parameter \texttt{verbose} to \texttt{TRUE}. The feature selection starts as follows:

\begin{verbatim}
R> result.search.fs <- ts(wine, 'V14', wrapper, iter=10, tamTabuList=4,
+   intensification=1, iterIntensification=5, diversification=1,
+   iterDiversification=5, verbose=TRUE)
\end{verbatim}
\begin{verbatim}
TS | InitialVector=0110100110010 | InitialFitness=0.8259
TS | Iter=1 | Vector=1110100110010 | Fitness=0.9163 | BestFitness=0.9163
TS | Iter=2 | Vector=1110100110011 | Fitness=0.9378 | BestFitness=0.9378
TS | Iter=3 | Vector=1110100111011 | Fitness=0.9719 | BestFitness=0.9719
TS | Iter=4 | Vector=1100100111011 | Fitness=0.9889 | BestFitness=0.9889
TS | Iter=5 | Vector=1100110111011 | Fitness=0.9833 | BestFitness=0.9889
TS | Iter=6 | Vector=1100111111011 | Fitness=0.9784 | BestFitness=0.9889
TS | Iter=7 | Vector=1100101111011 | Fitness=0.9892 | BestFitness=0.9892
TS | Iter=8 | Vector=1100101111001 | Fitness=0.9833 | BestFitness=0.9892
TS | Iter=9 | Vector=1100111111001 | Fitness=0.9889 | BestFitness=0.9892
TS | Iter=10 | Vector=1100111111101 | Fitness=0.9778 | BestFitness=0.9892
TS | Intensification stage  1
TS | InitialVector=1100111111001 | InitialFitness=0.9886
TS | Iter=1 | Vector=1100101111001 | Fitness=0.9830 | BestFitness=0.9886
TS | Iter=2 | Vector=1100101011001 | Fitness=0.9889 | BestFitness=0.9889
TS | Iter=3 | Vector=1100101011011 | Fitness=0.9778 | BestFitness=0.9889
TS | Iter=4 | Vector=1100111011011 | Fitness=0.9941 | BestFitness=0.9941
TS | Iter=5 | Vector=1100111011001 | Fitness=0.9833 | BestFitness=0.9941
TS | Diversification stage  1
TS | InitialVector=0011001000100 | InitialFitness=0.9163
TS | Iter=1 | Vector=1011001000100 | Fitness=0.9836 | BestFitness=0.9836
TS | Iter=2 | Vector=1011001000101 | Fitness=0.9889 | BestFitness=0.9889
TS | Iter=3 | Vector=1011001001101 | Fitness=0.9941 | BestFitness=0.9941
TS | Iter=4 | Vector=1011001001111 | Fitness=0.9947 | BestFitness=0.9947
TS | Iter=5 | Vector=1111001001111 | Fitness=0.9941 | BestFitness=0.9947
\end{verbatim}

After performing the feature selection, we can see the results by evaluating (running) the output object of the previous method:

\begin{verbatim}
R> result.search.fs
\end{verbatim}
\begin{verbatim}
$bestFeatures
     V1 V2 V3 V4 V5 V6 V7 V8 V9 V10 V11 V12 V13
[1,]  1  0  1  1  0  0  1  0  0   1   1   1   1

$bestFitness
[1] 0.9947

$basicStage
$basicStage$bestNeighbor
$basicStage$bestNeighbor$IterInitial
                V1 V2 V3 V4 V5 V6 V7 V8 V9 V10 V11 V12 V13   Fitness
neighborInitial  0  1  1  0  1  0  0  1  1   0   0   1   0    0.8259

$basicStage$bestNeighbor$Iter1
             V1 V2 V3 V4 V5 V6 V7 V8 V9 V10 V11 V12 V13   Fitness
bestNeighbor  1  1  1  0  1  0  0  1  1   0   0   1   0    0.9163

$basicStage$bestNeighbor$Iter2
             V1 V2 V3 V4 V5 V6 V7 V8 V9 V10 V11 V12 V13   Fitness
bestNeighbor  1  1  1  0  1  0  0  1  1   0   0   1   1    0.9378

$basicStage$bestNeighbor$Iter3
             V1 V2 V3 V4 V5 V6 V7 V8 V9 V10 V11 V12 V13   Fitness
bestNeighbor  1  1  1  0  1  0  0  1  1   1   0   1   1    0.9719

$basicStage$bestNeighbor$Iter4
             V1 V2 V3 V4 V5 V6 V7 V8 V9 V10 V11 V12 V13   Fitness
bestNeighbor  1  1  0  0  1  0  0  1  1   1   0   1   1    0.9889

$basicStage$bestNeighbor$Iter5
             V1 V2 V3 V4 V5 V6 V7 V8 V9 V10 V11 V12 V13   Fitness
bestNeighbor  1  1  0  0  1  1  0  1  1   1   0   1   1    0.9833

...

$basicStage$tabuList
$basicStage$tabuList$IterInitial
             V1 V2 V3 V4 V5 V6 V7 V8 V9 V10 V11 V12 V13
4IterToLeave  0  1  1  0  1  0  0  1  1   0   0   1   0
3IterToLeave NA NA NA NA NA NA NA NA NA  NA  NA  NA  NA
2IterToLeave NA NA NA NA NA NA NA NA NA  NA  NA  NA  NA
1IterToLeave NA NA NA NA NA NA NA NA NA  NA  NA  NA  NA

$basicStage$tabuList$Iter1
             V1 V2 V3 V4 V5 V6 V7 V8 V9 V10 V11 V12 V13
4IterToLeave  1  1  1  0  1  0  0  1  1   0   0   1   0
3IterToLeave  0  1  1  0  1  0  0  1  1   0   0   1   0
2IterToLeave NA NA NA NA NA NA NA NA NA  NA  NA  NA  NA
1IterToLeave NA NA NA NA NA NA NA NA NA  NA  NA  NA  NA

$basicStage$tabuList$Iter2
             V1 V2 V3 V4 V5 V6 V7 V8 V9 V10 V11 V12 V13
4IterToLeave  1  1  1  0  1  0  0  1  1   0   0   1   1
3IterToLeave  1  1  1  0  1  0  0  1  1   0   0   1   0
2IterToLeave  0  1  1  0  1  0  0  1  1   0   0   1   0
1IterToLeave NA NA NA NA NA NA NA NA NA  NA  NA  NA  NA

$basicStage$tabuList$Iter3
             V1 V2 V3 V4 V5 V6 V7 V8 V9 V10 V11 V12 V13
4IterToLeave  1  1  1  0  1  0  0  1  1   1   0   1   1
3IterToLeave  1  1  1  0  1  0  0  1  1   0   0   1   1
2IterToLeave  1  1  1  0  1  0  0  1  1   0   0   1   0
1IterToLeave  0  1  1  0  1  0  0  1  1   0   0   1   0

$basicStage$tabuList$Iter4
             V1 V2 V3 V4 V5 V6 V7 V8 V9 V10 V11 V12 V13
4IterToLeave  1  1  0  0  1  0  0  1  1   1   0   1   1
3IterToLeave  1  1  1  0  1  0  0  1  1   1   0   1   1
2IterToLeave  1  1  1  0  1  0  0  1  1   0   0   1   1
1IterToLeave  1  1  1  0  1  0  0  1  1   0   0   1   0

$basicStage$tabuList$Iter5
             V1 V2 V3 V4 V5 V6 V7 V8 V9 V10 V11 V12 V13
4IterToLeave  1  1  0  0  1  1  0  1  1   1   0   1   1
3IterToLeave  1  1  0  0  1  0  0  1  1   1   0   1   1
2IterToLeave  1  1  1  0  1  0  0  1  1   1   0   1   1
1IterToLeave  1  1  1  0  1  0  0  1  1   0   0   1   1

...
\end{verbatim}

The output is a list that shows the following elements: 1) \texttt{\$bestFeatures}: the best subset of features found, where the features selected by the procedure are marked with 1, 2) \texttt{\$bestFitness}: the fitness or error associated to the previous subset, 3) \texttt{\$basicStage}: the best subset of features found at each iteration (\texttt{\$bestNeighbor}), and the status of the taboo list in each of the iterations (\texttt{\$tabuList}), 4) \texttt{\$intensificationStage1}: the best subset of features found in each iteration in the intensification phase (\texttt{\$bestNeighbor}), and the status of the taboo list in each of the iterations in the intensification phase (\texttt{\$tabuList}), 5) \texttt{\$diversificationStage1}: the best subset of features found in each iteration in the diversification phase (\texttt{\$bestNeighbor}), and the status of the taboo list in each of the iterations in the diversification phase (\texttt{\$tabuList}). For the sake of brevity, steps 4 and 5 have been omitted since their structure is the same as the one for step 3. Complete results can be consulted at the package website.

\subsection{Filter approach}

The example below shows a feature selection on the \texttt{iris} dataset using a filter measure along with a genetic algorithm. The procedure is simpler than for wrapper measures since we do not have to generate a wrapper object.

The procedure consists of defining the search strategy and selecting the filter measure to be used as an evaluation, and then proceeding with the feature selection. As in the example above, search strategies can be used with default parameters. In this example we use a genetic algorithm, \texttt{ga()}, with parameters: 1) \texttt{iris} dataset, 2) ``Species'', 3) the \texttt{featureSetEval} parameter is the filter measure with default parameters, \texttt{giniIndex()}. The call to the function is as follows:

\begin{verbatim}
R> result.search.fs <- ga(iris, 'Species', giniIndex)
\end{verbatim}

In this example we change some of the default parameters of the genetic algorithm. The population size, \texttt{popSize}, is set to 10. The probability of crossing and mutation, \texttt{pcrossover} and \texttt{pmutation}, are set to 0.8 and 0.1, respectively. The number of iterations of the algorithm is set to 5. And finally to show the status of the execution per console the \texttt{verbose} parameter is set to \texttt{TRUE}.

\begin{verbatim}
R> result.search.fs <- ga(iris, 'Species', giniIndex, popSize = 10,
+   pcrossover = 0.8, pmutation = 0.1, maxiter=5, verbose=TRUE)
\end{verbatim}
\begin{verbatim}
GA | iter = 1 | Mean = 0.7896362 | Best = 1.0000000
GA | iter = 2 | Mean = 0.8951741 | Best = 1.0000000
GA | iter = 3 | Mean = 0.8416148 | Best = 1.0000000
GA | iter = 4 | Mean = 0.8735556 | Best = 1.0000000
GA | iter = 5 | Mean = 0.9633926 | Best = 1.0000000
\end{verbatim}

After performing the feature selection, we can see the results by executing the output object:

\begin{verbatim}
R> result.search.fs
\end{verbatim}
\begin{verbatim}
$bestFeatures
     Sepal.Length Sepal.Width Petal.Length Petal.Width
[1,]            0           1            1           1
[2,]            1           1            1           0
[3,]            1           1            1           1

$bestFitness
[1] 1

$population
      x1 x2 x3 x4   fitness
 [1,]  0  1  1  0 0.9844444
 [2,]  0  1  1  1 1.0000000
 [3,]  0  1  1  1 1.0000000
 [4,]  1  0  0  0 0.6805926
 [5,]  0  1  1  0 0.9844444
 [6,]  1  1  1  0 1.0000000
 [7,]  1  1  1  0 1.0000000
 [8,]  0  1  1  1 1.0000000
 [9,]  1  1  1  1 1.0000000
[10,]  0  1  1  0 0.9844444
\end{verbatim}

The output is again a list containing the following elements: 1) \texttt{\$bestFeatures}: the result of feature selection, which contains the subsets of features that achieve the best evaluation measure, 2) \texttt{\$bestFitness}: the fitness achieved by the best subset of features found, and 3) \texttt{\$population}: the population in the final iteration.

In addition, both filter and wrapper methods can be used with direct selection methods. These methods return a subset of features according to some cutoff criteria. The following example shows a direct selection of the 2 best features of the \texttt{iris} dataset based on the evaluation measure returned by the filter method \texttt{gainRatio}.

\begin{verbatim}
R> selectKBest(iris, 'Species', gainRatio, k=2)
\end{verbatim}

\section{Other FS packages available on CRAN}

In this section, we enumerate and develop a short review of the main CRAN repository packages implementing feature selection methods. As commented at section~\ref{sec:intro}, most of the available packages do not address the problem in depth. These packages are either not fully dedicated, or implement only one specific method. Because of this reason we focus primarily on two packages that stand out from the others. We compare these packages with the \textbf{FSinR} package and highlight the main differences.

\paragraph{MXM}

The package \textbf{MXM} \cite{MXM17} contains several implementations of feature selection algorithms with the objective of providing one or more minimum subsets of features. The main algorithm is the SES (Statistically Equivalent Signatures). It is a constraint-based algorithm that returns several subsets of minimum-size features that statistically contain a similar predictive level. This algorithm extends the MMPC (Max-Min Parents and Children) algorithm. The package also implements a series of conditional independence tests, which are the basis on which these constraint-based algorithms are built. Other methods of feature selection contained are MMMB, FBED, IAMB, forward and backward regression. It contains code written in \texttt{C++} and has the capability to perform parallel computing. It also contains regularization methods such as Ridge and Lasso. The implementation of all these methods in the package is very flexible allowing for quite varied ways of use. In comparison to the \textbf{FSinR} package, \textbf{MXM} does not include the filter and wrapper methods most widely considered in the literature and used in practice. The package does not contain either a variety of search methods that offer versatility when used in combination with evaluation methods as FSinR does. Instead, the package implements specific methods for feature selection originally proposed by the package authors.

\paragraph{FSelector}

The \textbf{FSelector} package \cite{FSelector18} implements various filter and cutoff methods as well as various search functions which work with a wrapper function. The filter and search methods in the package are widely known in the literature. Among them, relief or chi-squared can be highlighted as filter methods and hill-climbing or backward selection as a search methods. The search methods have as a parameter a wrapper function that evaluates the subsets but this function is not implemented in the package. \textbf{FSelector} has major differences with respect to the package \textbf{FSinR}. With regard to filter methods, \textbf{FSelector} provides an interesting list of them, but not the possibility to integrate them into a search strategy as \textbf{FSinR} does. It only gives the possibility of using them directly. On the other hand, the search methods are basic and the package does not offer more elaborated and powerful search strategies. The number of filter and search methods is smaller than in FSinR. In addition, the methods only return the selected features. They are much less informative than those included in \textbf{FSinR}, since they do not show anything about the result of the evaluation measure or the selection process. Finally, the main drawback of the package is that the user has to implement his own wrapper method to evaluate the features, which complicates the use of the package. In contrast, the \textbf{FSinR} package allows an ease use of the hundreds of methods available in the \textbf{caret} package.

\paragraph{Other packages}

Other CRAN packages that address the problem of feature selection do so to a lesser extent, as they are not fully dedicated, or only provide a particular method or algorithm. Various of them contain only one algorithm implemented specifically for feature selection, such as the \textbf{Boruta} package that includes a wrapper algorithm, called \textit{Boruta}, built on top of a classifier based on Random Forest. This is also the case of the \textbf{spSFR} package that performs a feature selection and ranking by simultaneous perturbation stochastic approximation algorithm, as well as the \textbf{EFS} package which presents an ensemble of some feature selection methods whose combination is used to determine the importance of features. Some other CRAN packages use trees to find the features that best explain the target variable. The \textbf{varSelRF} package are based on Random Forest and uses backwards variable elimination and selection based on the importance spectrum, whereas the \textbf{featurefinder} package analyzes the residues of classification and regression trees. There are also packages that implement a genetic algorithm for FS; \textbf{gaselect} \cite{gaselect19}, \textbf{GenAlgo} \cite{GenAlgo18}, \textbf{mogavs} \cite{mogavs18}, \textbf{kofnGA} \cite{kofsGA15}. In addition, there are R packages that are not dedicated to feature selection specifically, but contain some functions to perform it, such as packages \textbf{caret} \cite{Caret08}, \textbf{CORElearn}, \textbf{RWeka} or \textbf{mlr}.

\section{FSinR highlights}

The \textbf{FSinR} package offers to the R community tools to conveniently carry out the task of feature selection. The package aims to include a large and diverse set of filter and wrapper methods widely referred to in the literature and used in the Machine Learning, Data Mining or Data Science tasks. This is complemented with well-known search methods in which filter and wrapper measures can be integrated. The package has been designed to be as intuitive to use as possible. For example, the call to the package methods follows a uniform structure, which makes the package have a homogeneous user interface. Then if you know how to call a function you know how to call any other function. All the above makes the package easy to extend and update in the future. In addition, the package allows the use of all the models available in \textbf{caret} as wrapper measures. Furthermore, the package can handle all types of data (numerical, categorical,...), problems of diverse nature (regression or classification) and use different types of models to generate evaluation measures (Random Forest, Neural Networks, SVM,...). The package checks internally that the types of the passed parameters match the expected ones, making its use more robust. The reasons detailed above make \textbf{FSinR} an FS package apart from the currently available ones. We intend to keep extending \textbf{FSinR} functionally by adding new methods filter and search methods in future releases.

\section{Conclusion}

This paper presents the \textbf{FSinR} package for feature selection. It implements the most commonly used filter and wrapper methods in the literature as well as search methods. More specifically, it implements 16 filter methods and can use as wrapper methods all the models available in caret for classification and regression tasks, currently 238. These filter and wrapper methods can be used directly or combined with the 13 search algorithms included in the package to carry out the feature selection process. Some examples of the use of the package and a comparison with other CRAN packages with the same purpose have been provided. The main highlights of \textbf{FSinR} are its robustness, its extension and extensibility, and the versatility of use offered. All these properties make \textbf{FSinR} a package apart in the Feature Selection topic. 

\section{Acknowledgments}

This research has been partially funded by the Spanish National Research Projects TIN2013-47210-P, and TIN2016-81113-R and by the Andalusian Regional Government Excellence Research Project P12-TIC-2958.

\bibliographystyle{unsrt}  
\bibliography{references}  

\end{document}